\documentclass[runningheads]{llncs}
\usepackage{graphicx}

\usepackage{tikz}
\usepackage{comment}
\usepackage{amsmath,amssymb} 
\usepackage{color}

\usepackage{todonotes}
\usepackage{amsmath}
\usepackage{dblfloatfix}

\usepackage{graphicx}
\usepackage{subcaption}
\usepackage{tikz}
\usetikzlibrary{decorations.pathreplacing}

\begin{document}
\pagestyle{headings}
\mainmatter
\def\ECCVSubNumber{17}  

\title{Real-Time Sign Language Detection using Human Pose Estimation} 

\authorrunning{Moryossef A., Tsochantaridis I., Aharoni R., Ebling S. and Narayanan S.} 


\titlerunning{Real-Time Sign Language Detection using Human Pose Estimation}

\author{Amit Moryossef\inst{1,2} \and
Ioannis Tsochantaridis\inst{1} \and
Roee Aharoni\inst{1} \and \\
Sarah Ebling\inst{3} \and
Srini Narayanan\inst{1}}
%
%
\institute{Google \email{\{ioannis,roeeaharoni,srinin\}@google.com} \and
Bar-Ilan University \email{amitmoryossef@gmail.com} \and
University of Zurich \email{ebling@cl.uzh.ch}}

\maketitle
\setcounter{footnote}{0}

\begin{abstract}
We propose a lightweight real-time sign language detection model, as we identify the need for such a case in videoconferencing.
We extract optical flow features based on human pose estimation and, using a linear classifier, show these features are meaningful with an accuracy of 80\%, evaluated on the Public DGS Corpus.
Using a recurrent model directly on the input, we see improvements of up to 91\% accuracy, while still working under 4ms.
We describe a demo application to sign language detection in the browser in order to demonstrate its usage possibility in videoconferencing applications.
\end{abstract}

\section{Introduction}\label{sec:intro}
Sign language detection \cite{borg2019sign} is defined as the binary-classification task for any given frame of a video if a person is using sign-language or not. 
Unlike sign language recognition \cite{cihan2018neural,cui2017recurrent}, where the task is to recognize the form and meaning of signs in a video, or sign language identification, where the task is to identify \emph{which} sign language is used, the task of sign language detection is to detect \emph{when} something is being signed.


With the recent rise of videoconferencing platforms, we identify the problem of signers not ``getting the floor'' when communicating, which either leads to them being ignored or to a cognitive load on other participants, always checking to see if someone starts signing.
Hence, we focus on the real-time sign language detection task with uni-directional context to allow for videoconferencing sign language prominence.

We propose a simple human optical-flow representation for videos based on pose estimation (\S\ref{sec:model:rep}), which is fed to a temporally sensitive neural network (\S\ref{sec:model:net}) to perform a binary classification per frame --- is the person signing or not.
We compare various possible inputs, such as full-body pose estimation, partial pose estimation, and bounding boxes (\S\ref{sec:experiments}), and contrast their acquisition time in light of our targeted real-time application.

We demonstrate our approach on the Public DGS Corpus (German Sign Language) \cite{hanke-etal-2020-extending}, using full-body pose estimation \cite{schulderopenpose} collected through OpenPose \cite{openpose,simon2017hand}.
We show results of 87\%-91\% prediction accuracy depending on the input, with per-frame inference time of $350-3500\mu$s (\S\ref{sec:results}),
and release our training code and models\footnote{\url{\scriptsize{https://github.com/google-research/google-research/tree/master/sign\_language\_detection}}}.

\section{Background}\label{sec:background}
The computational sign language processing (SLP) literature rarely addresses detection \cite{borg2019sign} and mainly focuses on sign language recognition \cite{cihan2018neural,cui2017recurrent,konstantinidis2018sign} and identification \cite{gebre2013automatic,monteiro2016detecting}.

\subsection{Sign Language Detection}
Previous work \cite{borg2019sign} introduces the classification of frames taken from YouTube videos as either signing or not. They take a spatial and temporal approach based on VGG-16 CNN \cite{simonyan2014very} to encode each frame and use a GRU \cite{cho2014learning} to encode the sequence of frames, in a window of 20 frames at 5fps. In addition to the raw frame, they also either encode optical flow history, aggregated motion history, or frame difference.
However, for our use case, 5fps might not be enough, as it introduces an artificial 200ms delay from when a person starts signing to when they could be detected.
Furthermore, this network takes upwards of 3 seconds to run on CPU per inference.

Most recently, Apple \cite{apple2020wwdc} announced sign language detection for group FaceTime calls in iOS 14, iPadOS 14, and macOS Big Sur. 
They did not share any implementation details of their detection model, which makes it hard to compare their model to the one we propose in this paper.
Nonetheless, as FaceTime group calls are encrypted end-to-end, we assume that the detection happens on-device rather than on the server-side.

\subsection{Sign Language Recognition}
Sign language recognition has been widely studied across different domains and sign languages. As sign language corpora are usually small \cite{bragg2019sign}, previous works take one of two approaches to reduce the network's parameters: (1) using pose estimation on the original videos \cite{isaacs2004hand,zafrulla2011american,konstantinidis2018sign}; or (2) using pre-trained CNNs to get a feature vector per frame \cite{cui2017recurrent,cihan2018neural}.
While different, both methods can encode adequate features to be used for recognition. 
Studies of human signers have shown that detailed information like exact descriptions of the hand shape are not always required for humans to interpret sign language \cite{poizner1981perception,sperling1985intelligible}.

Looking at examples of sign videos, we hypothesize that the most challenging part of this task is to identify when a person starts signing, because a signer might initiate hand movement for other purposes, for example, to touch their face. Distinguishing this type of ambient motion from actual linguistic sign movement is not always straightforward. Although not explicitly studied on signers, studies find the average person touches their face between 15.7 and 23 times per hour \cite{nicas2008study,kwok2015face}. Further complicating this issue, people in different cultures exhibit different face-touching patterns, including frequency, area, and hand preference \cite{hatta1984differences}.

\subsection{Sign Language Identification}
A study \cite{gebre2013automatic} finds that a random-forest classifier can distinguish between British Sign Language (BSL) and Greek Sign Language (ENN) with a 95\% F1 score.
This finding is further supported by more recent work \cite{monteiro2016detecting} which manages to differentiate between British Sign Language and French Sign Language (Langue des Signes Fran\c{c}aise, LSF) with 98\% F1 score in videos with static backgrounds, and between American Sign Language and British Sign Language with 70\% F1 score for videos mined from popular video sharing sites. The authors attribute their success mainly to the different fingerspelling systems, which is two-handed in the case of BSL and one-handed in the case of ASL and LSF.

\section{Model}\label{sec:model}
For a video, for every frame given, we would like to predict whether the person in the video is signing or not.

\subsection{Input Representation}\label{sec:model:rep}
As evident by previous work \cite{borg2019sign}, using the raw frames as input is computationally expensive, and noisy.
Alternatively, in computer vision, optical flow is one way to calculate the movement of every object in a scene. However, because signing is inherently a human action, we do not care about the flow of every object, but rather only the flow of the human. Optimally, we would like to track the movement of every pixel on the human body from one frame to another, to gauge its movement vector.
As a proxy to such data, we opt for full-body human pose estimation, defining a set of points detected in every video frame that marks informative landmarks, like joints and other moving parts (mouth, eyes, eyebrows, and others).

Getting the optical flow $F$ for these predefined points $P$ at time $t$ is then well defined as the L2 norm of the vector resulting from subtracting every two consecutive frames. We normalize the flow by the frame-rate in which the video was captured for the representation to be frame-rate invariant (Equation \ref{eq:rep}).
\begin{equation}\label{eq:rep}
    F(P)_t = ||P_t - P_{t-1}||_{2} * fps
\end{equation} 

We note that if a point $p$ was not identified in a given frame $t$, the value of $F(p)_t$ and $F(p)_{t+1}$ automatically equals to 0. 
This is done to avoid introducing fake movements from a poor pose estimation system or unknown movement from landmarks going out-of-frame.

An additional benefit of using full-body pose estimation is that we can normalize the size of all people, regardless of whether they use a high-/low-resolution camera and the distance at which they are from the camera.

\begin{figure*}[ht]
    \begin{subfigure}{0.9\linewidth}
        \hspace{0.3em}
        \includegraphics[width=\linewidth,height=4em]{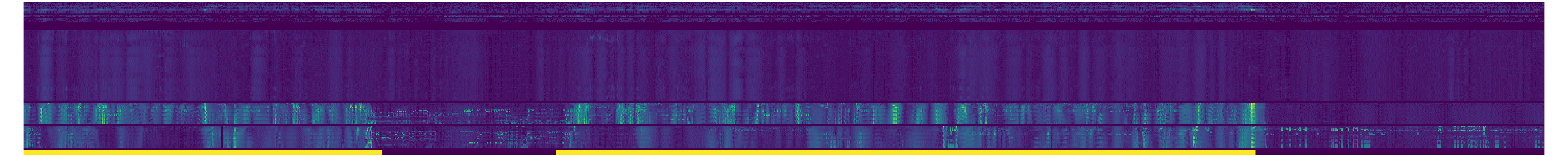}
        \label{fig:rep:1}
    \end{subfigure}
    \begin{subfigure}{0.09\linewidth}
        \vspace{-1.2em}\hspace{-1em}
        \begin{tikzpicture}
            \draw [decorate,decoration={brace,amplitude=3pt,mirror,raise=3pt},yshift=0pt]
            (0,1.073) -- (0,1.266) node [black,midway,xshift=0.55cm] {\tiny body };
            
            \draw [decorate,decoration={brace,amplitude=3pt,mirror,raise=3pt},yshift=0pt]
            (0,0.473) -- (0,1.06) node [black,midway,xshift=0.5cm] {\tiny face};
            
            \draw [decorate,decoration={brace,amplitude=3pt,mirror,raise=3pt},yshift=0pt]
            (0,0.273) -- (0,0.46) node [black,midway,xshift=0.5cm] {\tiny left};
            
            \draw [decorate,decoration={brace,amplitude=3pt,mirror,raise=3pt},yshift=0pt]
            (0,0.073) -- (0,0.26) node [black,midway,xshift=0.58cm] {\tiny right};
        \end{tikzpicture}
    \end{subfigure}
    \begin{subfigure}{0.9\linewidth}
        \vspace{-0.5em}\hspace{0.3em}
        \includegraphics[width=\linewidth,height=4em]{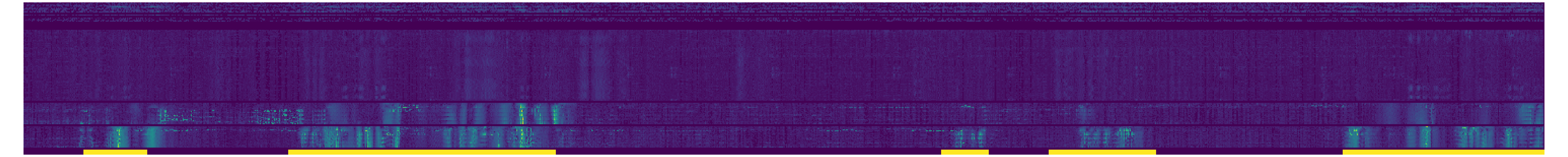}
        \label{fig:rep:2}
    \end{subfigure}
    \begin{subfigure}{0.09\linewidth}
        \vspace{-0.5em}
        \vspace{-1.2em}\hspace{-1em}
        \begin{tikzpicture}
            \draw [decorate,decoration={brace,amplitude=3pt,mirror,raise=3pt},yshift=0pt]
            (0,1.073) -- (0,1.266) node [black,midway,xshift=0.55cm] {\tiny body };
            
            \draw [decorate,decoration={brace,amplitude=3pt,mirror,raise=3pt},yshift=0pt]
            (0,0.473) -- (0,1.06) node [black,midway,xshift=0.5cm] {\tiny face};
            
            \draw [decorate,decoration={brace,amplitude=3pt,mirror,raise=3pt},yshift=0pt]
            (0,0.273) -- (0,0.46) node [black,midway,xshift=0.5cm] {\tiny left};
            
            \draw [decorate,decoration={brace,amplitude=3pt,mirror,raise=3pt},yshift=0pt]
            (0,0.073) -- (0,0.26) node [black,midway,xshift=0.58cm] {\tiny right};
        \end{tikzpicture}
    \end{subfigure}
    \vspace{-1.3em}
    \caption{
    Optical-flow norm representation of a conversation between two signers.  
    The x-axis is the progression of time, 1,500 frames over 30 seconds in total. 
    The yellow marks are the gold labels for spans when a signer is signing.
    }
    \label{fig:rep}
\end{figure*}
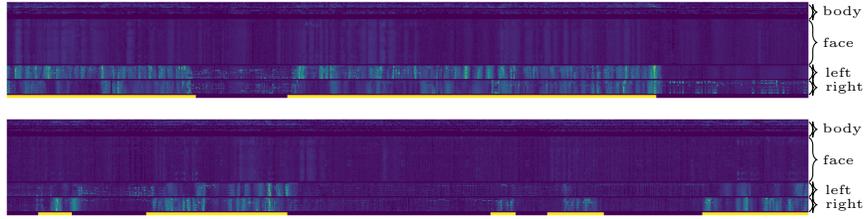

\subsection{Temporal Model}\label{sec:model:net}
Figure \ref{fig:rep} demonstrates our input representation for an example video. 
It shows, to the naked eye, that this representation is meaningful. The movement, indicated by the bright colors, is well aligned with the gold spans annotation. Thus, we opt to use a shallow sequence tagging model on top of it.

We use a uni-directional LSTM \cite{hochreiter1997long} with one layer and 64 hidden units directly on this input, normalized for frame rate, and project the output to a 2-dimensional vector.
For training, we use the negative-log-likelihood loss on the predicted classes for every frame.
For inference, we take the arg-max of the output vector (Equation \ref{eq:model}).

\begin{equation}\label{eq:model}
    signing(P) = \operatorname*{arg\,max} LSTM(F(P)) * W
\end{equation}

Note that this model allows us to process each frame as we get it, in real-time, by performing a single step of the LSTM and project its output.
Unlike autoregressive models, we do not feed the last-frame classification as input for the next frame, as just classifying the new frame with the same tag would almost get 100\% accuracy on this task, depending on gold labels to be available. Instead, we rely on the hidden state of the LSTM to hold such information as a probability.

\section{Experiments}\label{sec:experiments}

The Public DGS Corpus \cite{hanke-etal-2020-extending} includes 301 videos with an average duration of 9 minutes, of two signers in conversation\footnote{There are also monologue story-telling, but both signers are always shown.}, at 50fps. Each video includes gloss annotations and spoken language translations (German and English). Using this information, we mark each frame as either ``signing'' (50.9\% of the data) or ``not-signing'' (49.1\% of the data) depending on whether it belongs to a gloss segment.
Furthermore, this corpus is enriched with OpenPose \cite{openpose} full-body pose estimations \cite{schulderopenpose} including 137 points per frame (70 for the face, 25 for the body, and 21 for each hand). 
In order to disregard video resolution and distance from the camera, we normalize each of these poses such that the mean distance between the shoulders of each person equals 1.
We split this dataset into 50:25:25 for training, validation, and test, respectively. For every ``part'' (face, body, left and right hands), we also calculate its bounding box based on the minimum and maximum value of all of the landmarks.

\begin{figure*}[t]
    \begin{subfigure}{0.245\linewidth}
        \includegraphics[width=\linewidth]{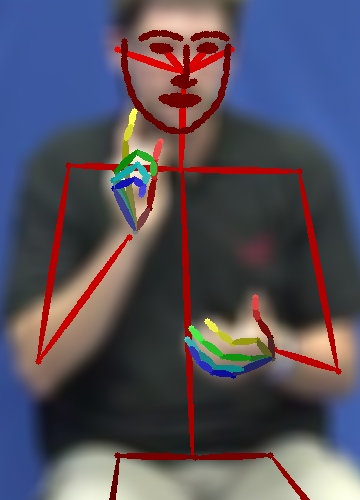}
        \caption{Pose-All}
        \label{fig:exp:1}
    \end{subfigure}
     \begin{subfigure}{0.245\linewidth}
        \includegraphics[width=\linewidth]{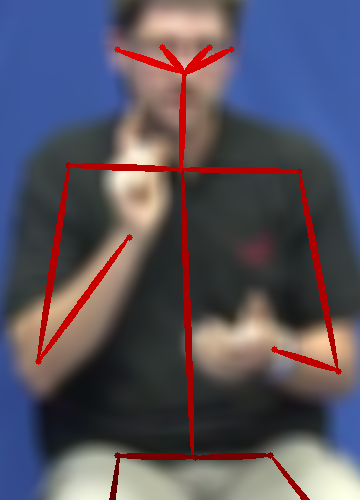}
        \caption{Pose-Body}
        \label{fig:exp:2}
    \end{subfigure}
    \begin{subfigure}{0.245\linewidth}
        \includegraphics[width=\linewidth]{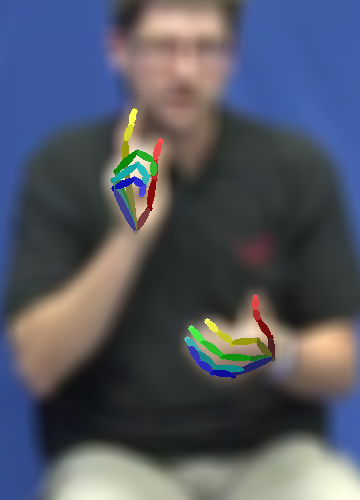}
        \caption{Pose-Hands}
        \label{fig:exp:3}
    \end{subfigure}
    \begin{subfigure}{0.245\linewidth}
        \includegraphics[width=\linewidth]{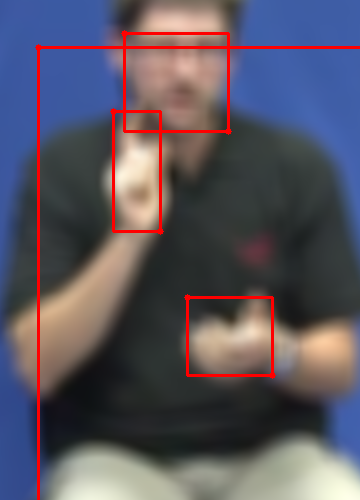}
        \caption{BBOX}
        \label{fig:exp:4}
    \end{subfigure}
    \caption{Visualization of our different experiments inputs.}
    \label{fig:exp}
\end{figure*}

We experiment with three linear baselines with a fixed context (Linear-1, Linear-25, Linear-50) and four experimental recurrent models with different counts of input features:
\begin{enumerate}
    \item \textbf{Pose-All}---including all of the landmarks from the poses. (f. \ref{fig:exp:1})
    \item \textbf{Pose-Body}---including only the body landmarks. (f. \ref{fig:exp:2})
    \item \textbf{Pose-Hands}---including only the left- and right-hand landmarks. (f. \ref{fig:exp:3})
    \item \textbf{BBOX}---including the bounding boxes of the face, body, and hands. (f. \ref{fig:exp:4})
\end{enumerate}

Finally, we measure the execution time of each model on CPU, using an Intel(R) Xeon(R) CPU E5-2650 v4 @ 2.20GHz.
We measure the execution time per frame given a single frame at a time, using multiple frameworks: Scikit-Learn (sk) \cite{scikit-learn}, TensorFlow (tf) \cite{tensorflow2015-whitepaper} and PyTorch (pt) \cite{pytorch}.

\section{Results}\label{sec:results}

Table \ref{table:results} includes the accuracy and inference times for each of our scenarios. 
Our baseline systems show that using a linear classifier with a fixed number of context frames achieves between 79.9\% to 84.3\% accuracy on the test set.
However, all of the baselines perform worse than our recurrent models, for which we achieve between 87.7\% to 91.5\% accuracy on the test set. Generally, we see that using more diverse sets of landmarks performs better. 
Although the hand landmarks are very indicative, using just the hand BBOX almost matches in accuracy, and using the entire body pose, with a single point per hand, performs much better.
Furthermore, we see that regardless of the number of landmarks used, our models generally perform faster the fewer landmarks are used. We note that the prediction time varies between the different frameworks, but does not vary much within a particular framework. It is clear, however, that the speed of these models' is sufficient, as even the slowest model, using the slowest framework, runs at 285 frames-per-second on CPU.

We note from manually observing the gold data that sometimes a gloss segment starts before the person actually begins signing, or moving at all. This means that our accuracy ceiling is not 100\%. We did not perform a rigorous re-annotation of the dataset to quantify how extensive this problem is.

\begin{table}[ht]
    \begin{center}
    \resizebox{.9\textwidth}{!}{
    \begin{tabular}{|l|l|l|l|l|l|l|l|}
        \hline
        \textbf{Model} & \textbf{Points} & \textbf{Params} & \textbf{Dev Acc} & \textbf{Test Acc} & \textbf{$\partial t$ (sk)} & \textbf{$\partial t$ (tf)} & \textbf{$\partial t$ (pt)} \\ \hline\hline
        Linear-1 & $25$ & $25$ & $79.99\%$ & $79.93\%$ & $6.49\mu$s & $823\mu$s & $2.75\mu$s \\ \hline
        Linear-25 & $25$ & $625$ & $84.13\%$ & $83.79\%$ & $6.78\mu$s & $824\mu$s & $5.10\mu$s \\ \hline
        Linear-50 & $25$ & $1,250$ & $85.06\%$ & $83.39\%$ & $6.90\mu$s & $821\mu$s & $7.41\mu$s \\ \hline\hline
        BBOX & $8$ & $18,818$ & $87.49\%$ & $87.78\%$ & --- & $3519\mu$s & $367\mu$s \\ \hline
        Pose-Hands & $42$ & $27,522$ & $87.65\%$ & $88.05\%$ & --- &  $3427\mu$s & $486\mu$s \\ \hline
        Pose-Body & $25$ & $23,170$ & $92.17\%$ & $90.35\%$ & --- &  $3437\mu$s & $443\mu$s \\ \hline
        Pose-All & $137$ & $51,842$ &  $92.31\%$ & $91.53\%$ & --- & $3537\mu$s & $588\mu$s \\ \hline
    \end{tabular}
    }
    \end{center}
    \caption{Accuracy and inference-time ($\partial t$) results for the various experiments.}
    \label{table:results}
\end{table}

\section{Analysis}\label{sec:analysis}
\begin{figure*}[b]
    \centering
    \includegraphics[keepaspectratio=true,width=0.3\linewidth]{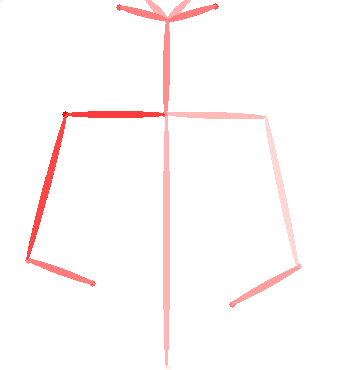}
    \caption{The average pose in the dataset. The opacity of every landmark is determined by its coeffient in the \emph{Linear-1} model.}
    \label{fig:avg_pose}
\end{figure*}

As we know that different pose landmarks have varying importance to the classification, we use the \emph{Linear-1} model's coefficients magnitude to visualize how the different landmarks contribute. Figure \ref{fig:avg_pose} visualizes the average human pose in the dataset, with the opacity of every landmark being the absolute value of the coefficient.

First, we note that the model attributes no importance to any landmark below the waist. This makes sense as they both do not appear in all videos, and bare no meaning in sign language. The eyes and nose seem to carry little weight, while the ears carry more. We do not attribute this to any signing phenomenon.

Additionally, we note hands asymmetry. While both wrists have a high weight, the elbow and shoulder for the right hand carry more weights than their corresponding left counterparts. This could be attributed to the fact that most people are right handed, and that in some sign languages the signer must decide which hand is dominant in a consistent manner. We see this asymmetry as a feature of our model, and note that apps using our models could also include a ``dominant hand'' selection.

\begin{figure*}[t]
    \centering
    \includegraphics[keepaspectratio=true,width=\linewidth]{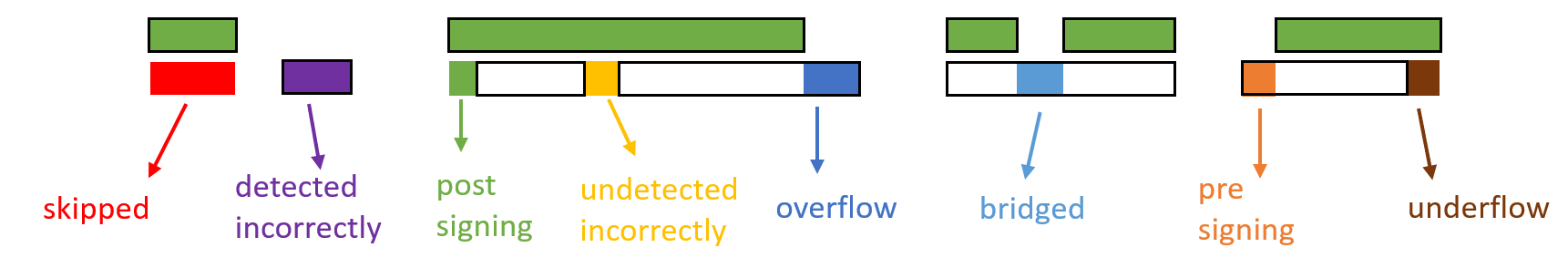}
    \caption{
    Visualization of the different types of errors. 
    The first row contains the gold annotations, and the second row contains a model's prediction.
    }
    \label{fig:errors}
\end{figure*}

To further understand what situations our models capture, we check multiple properties of them on the test set.
We start by generally noting that our data is conversational. 84.87\% of the time, only one participant is signing, while 8.5\% of the time both participants are signing, and in the remaining 6.63\% of the time no one is signing, primarily when the participants are being instructed on the task.

Our test set includes $4,138$ \emph{signing} sequences with an average length of $11.35$ seconds, and a standard deviation of $29.82$ seconds. 
It also includes $4,091$ \emph{not-signing} sequences with an average length of $9.95$ seconds, and a standard deviation of $24.18$ seconds. 

For each of our models, we compare the following error types (Figure \ref{fig:errors}):
\begin{itemize}
    \item \textbf{Bridged}---Cases where the model bridged between two signing sections, still predicting the person to be \emph{signing} while the annotation says they are not.
    \item \textbf{Signing Detected Incorrectly}---Cases where the model predicted a \emph{signing} span fully contained within a \emph{not-signing} annotation.
    \item \textbf{Signing Overflow}---Cases where signing was still predicted after a \emph{signing} section ended.
    \item \textbf{Started Pre-Signing}---Cases where \emph{signing} was predicted before a \emph{signing} section started.
    \item \textbf{Skipped}---Cases where the model did not detect entire \emph{signing} sections.
    \item \textbf{Signing Undetected Incorrectly}---Cases where the model predicted a \emph{not-signing} span fully contained within a \emph{signing} annotation.
    \item \textbf{Started Post-Signing}---Cases where the \emph{signing} section started before it was predicted to start.
    \item \textbf{Signing Underflow}---Cases where the \emph{signing} section was predicted to end prematurely.
\end{itemize}

\begin{table}[h]
    \resizebox{0.85\textwidth}{!}{
    \begin{tabular}{|l|l|l|l|l|l|l|l|}
        \hline
 & \textbf{linear-1} & \textbf{linear-25} & \textbf{linear-50} \\ \hline
Bridged & $ 107$ ($0.10\pm0.15$) & $ 308$ ($0.34\pm0.40$) & $ 426$ ($0.45\pm0.46$) \\ \hline
Signing Detected Incorrectly & $132151$ ($0.04\pm0.07$) & $8773$ ($0.30\pm0.81$) & $6594$ ($0.34\pm1.06$) \\ \hline
Signing Overflow & $4094$ ($0.09\pm0.15$) & $3893$ ($0.32\pm0.43$) & $3775$ ($0.46\pm1.17$) \\ \hline
Started Pre-Signing & $ 873$ ($0.09\pm0.13$) & $ 345$ ($0.45\pm0.68$) & $ 257$ ($0.88\pm4.27$) \\ \hline
Skipped & $  50$ ($1.41\pm1.95$) & $ 298$ ($1.38\pm1.43$) & $ 446$ ($1.49\pm1.60$) \\ \hline
Signing undetected incorrectly & $219531$ ($0.05\pm0.10$) & $26185$ ($0.27\pm0.50$) & $18037$ ($0.32\pm0.66$) \\ \hline
Started Post-Signing & $4199$ ($0.17\pm0.23$) & $3951$ ($0.48\pm0.57$) & $3803$ ($0.60\pm0.77$) \\ \hline
Signing Underflow & $1677$ ($0.15\pm0.26$) & $1092$ ($0.58\pm0.91$) & $ 827$ ($0.71\pm0.96$) \\ \hline
    \end{tabular}
    }
    \resizebox{\textwidth}{!}{
    \begin{tabular}{|l|l|l|l|l|l|l|l|}
        \hline
 & \textbf{BBOX} & \textbf{Pose-Hands} & \textbf{Pose-Body} & \textbf{Pose-All} \\ \hline
Bridged & $ 754$ ($0.97\pm1.94$) & $ 861$ ($1.26\pm2.63$) & $ 747$ ($1.12\pm2.35$) & $ 573$ ($0.75\pm1.08$) \\ \hline
Signing Detected Incorrectly & $5697$ ($0.64\pm1.93$) & $12919$ ($0.33\pm1.33$) & $6286$ ($0.38\pm1.29$) & $11384$ ($0.25\pm1.14$) \\ \hline
Signing Overflow & $3337$ ($0.95\pm2.10$) & $3230$ ($1.01\pm2.46$) & $3344$ ($0.67\pm1.29$) & $3518$ ($0.48\pm0.87$) \\ \hline
Started Pre-Signing & $ 402$ ($1.33\pm2.73$) & $ 558$ ($1.59\pm5.15$) & $ 298$ ($1.48\pm3.87$) & $ 408$ ($0.70\pm1.97$) \\ \hline
Skipped & $ 199$ ($1.31\pm1.40$) & $ 115$ ($1.45\pm1.54$) & $ 243$ ($1.31\pm1.30$) & $ 146$ ($1.41\pm1.42$) \\ \hline
Signing undetected incorrectly & $4089$ ($0.48\pm0.76$) & $3526$ ($0.26\pm0.51$) & $4786$ ($0.32\pm0.60$) & $5526$ ($0.23\pm0.44$) \\ \hline
Started Post-Signing & $3939$ ($0.34\pm0.44$) & $4023$ ($0.24\pm0.34$) & $3895$ ($0.37\pm0.49$) & $3992$ ($0.29\pm0.36$) \\ \hline
Signing Underflow & $ 370$ ($0.82\pm1.08$) & $ 297$ ($0.55\pm0.68$) & $ 506$ ($0.63\pm0.97$) & $ 666$ ($0.44\pm0.66$) \\ \hline
    \end{tabular}
    }
    \vspace{1em}
    \caption{We evaluate every model on the different error types, and show number of sequences with that error, including average sequence length in seconds and standard deviation.}
    \label{table:analysis}
\end{table}

Table \ref{table:analysis} includes the number of sequences, including average length and standard deviation in seconds, for each of the error types. Most notably, we see that the less context the model has, the more sporadic its predictions and thus it will generally completely bridge or skip less sequences. The same locality however introduces many signing detected / undetected incorrectly errors, albeit of short lengths.

In the sequential models, we generally see a lower number of sequences as they can incorporate global features in the classification.
As indicated by the accuracy scores, we see fewer errors of most types the more diverse the input points are, with one notable exception for the \emph{Pose-All} model which under-performs \emph{Pose-Body} on all errors except for \emph{Bridged} and \emph{Skipped}.

\section{Demo Application}\label{sec:demo}
With this publication, we release a demo application working in the browser for computers and mobile devices.
Pragmatically, we choose to use the ``Pose-Body'' model variant, as it performs almost on par with our best model, ``Pose-All'', and we find it is feasible to acquire the human body poses in real-time with currently available tools. 

We use PoseNet \cite{papandreou2017towards,papandreou2018personlab} running in the browser using TensorFlow.js \cite{smilkov2019tensorflow}.
PoseNet includes two main image encoding variants: MobileNet \cite{howard2017mobilenets}, which is a lightweight model aimed at mobile devices, and ResNet \cite{he2016deep}, which is a larger model that requires a dedicated GPU.
Each model includes many sub-variants with different image resolution and convolutional strides, to further allow for tailoring the network to the user's needs.
In our demo, we first tailor a network to the current computation device to run at least at 25fps.
While using a more lightweight network might be faster, it might also introduce pose estimation errors.

The pose estimation we use only returns 17 points compared to the 25 of  OpenPose; hence, we map the 17 points to the corresponding indexes for OpenPose.
We then normalize the body pose vector by the mean shoulder width the person had in the past 50 frames in order to disregard camera resolution and distance of the signer from the camera.

Onward, there are two options: either send the pose vector to the videoconferencing server where inference could be done or perform the inference locally. As our method is faster than real-time, we chose the latter and perform inference on the device using TensorFlow.js.
For every frame, we get a signing probability, which we then show on the screen.

In a production videoconferencing application, this signing probability should be streamed to the call server, where further processing could be done to show the correct people on screen. We suggest using the signing probability as a normalized ``volume'', such that further processing is comparable to videoconferencing users using speech.

While this is the recommended way to add sign language detection to a videoconferencing app, as the goal of this work is to empower signers, our demo application can trigger the speaker detection by transmitting audio when the user is signing. Transmitting ultrasonic audio at 20KHz, which is inaudible for humans, manages to fool Google Meet, Zoom and Slack into thinking the user is speaking, while still being inaudible. One limitation of this method is that videoconferencing app developers can crop the audio to be in the audible human range and thus render this application useless. Another limitation is that using high-frequency audio can sound crackly when compressed, depending on the signer's internet connection strength.

Our model and demo, in their current forms, only allow for the detection of a single signer per video stream. 
However, if we can detect more than a single person, and track which poses belong to which person in every frame, there is no limitation to run our model independently on each signer.

\section{Discussion}
\subsection{Limitations}\label{sec:limitations}
We note several limitations to our approach.
The first is that it relies on the pose estimation system to run in real-time on any user's device. This proves to be challenging, as even performing state-of-the-art pose estimation on a single frame on a GPU with OpenPose \cite{openpose,cao2017realtime} can take upwards of 300ms, which introduces two issues: (1) If in order to get the optical-flow, we need to pose two frames, we create a delay from when a person starts signing to when they could be accurately detected as signing, equal to at least two times the pose processing time. (2) Running this on mobile devices or devices without hardware acceleration like a GPU may be too slow.

As we only look at the input's optical flow norm, our model might not be able to pick up on times when a person is just gesturing rather than signing. However, as this approach is targeted directly at sign language users rather than the general non-signing public, erring on the side of caution and detecting any meaningful movements is preferred.

\subsection{Demographic Biases}
The data we use for training was collected from various regions of Germany, with equal number of males and females, as well as an equal number of participants from different age groups \cite{dgs_DataStatement}.
Although most of the people in the dataset are European white, we do not attribute any significance between the color of their skin to the performance of the system, as long as the pose estimation system is not biased.

Regardless of age, gender, and race, we do not address general ethnic biases such as different communities of signers outside of Germany signing differently - whether it is the size, volume, speed, or other properties.

\section{Conclusions}\label{sec:conclusions}

We propose a simple human optical-flow representation for videos based on pose estimation to perform a binary classification per frame --- is the person signing or not.
We compare various possible inputs, such as full-body pose estimation, partial pose estimation, and bounding boxes and contrast their acquisition time in light of our targeted real-time videoconferencing sign language detection application.

We demonstrate our approach on the Public DGS Corpus (German Sign Language), and show results of 87\%-91\% prediction accuracy depending on the input, with per-frame inference time of $350-3500\mu$s. 

\clearpage
%
%
\bibliographystyle{splncs04}
\bibliography{bib}
\end{document}